%% file: main.tex
\documentclass[10pt,twocolumn,letterpaper]{article}

\usepackage{cvpr}      % To produce the REVIEW version
\usepackage{algorithm}
\usepackage{algpseudocode}
\usepackage{multirow}
\usepackage{xcolor}
\usepackage{array}   % 推荐同时引入，用于列格式控制
\usepackage{booktabs} % 推荐同时引入，用于美化表格线条
\input{preamble}

\definecolor{cvprblue}{rgb}{0.21,0.49,0.74}
\usepackage[pagebackref,breaklinks,colorlinks,allcolors=cvprblue]{hyperref}

\def\paperID{*****} % *** Enter the Paper ID here
\def\confName{CVPR}
\def\confYear{****}

\title{CC-FMO: Camera-Conditioned Zero-Shot Single Image to 3D Scene Generation with Foundation Model Orchestration}

\author{Boshi Tang\\
Tsinghua University\\
Beijing, China\\
{\tt\small tbs16@mails.tsinghua.edu.cn}
\and
Henry Zheng\\
Tsinghua University\\
Beijing, China\\
{\tt\small jh-zheng22@mails.tsinghua.edu.cn}
\and
Rui Huang\\
Tsinghua University\\
Beijing, China\\
{\tt\small hr20@mails.tsinghua.edu.cn}
\and
Gao Huang\thanks{Corresponding author.}\\
Tsinghua University\\
Beijing, China\\
{\tt\small gaohuang@tsinghua.edu.cn}
}

\begin{document}
\maketitle
\input{sec/0_abstract}    
\input{sec/1_intro}
\input{sec/2_related_works}
\input{sec/3_methods}
\input{sec/4_experimentation}
\input{sec/5_Conclusion}
{
    \small
    \bibliographystyle{ieeenat_fullname}
    \bibliography{main}
}

\input{sec/X_suppl}

\end{document}

%% file: preamble.tex
\newcommand{\name}{CC-FMO}

%% file: sec/0_abstract.tex
\begin{abstract}

High-quality 3D scene generation from a single image is crucial for AR/VR and embodied AI applications. Early approaches struggle to generalize due to reliance on specialized models trained on curated small datasets. While recent advancements in large-scale 3D foundation models have significantly enhanced instance-level generation, coherent scene generation remains a challenge, where performance is limited by inaccurate per-object pose estimations and spatial inconsistency. To this end, this paper introduces \name{}, a zero-shot, camera-conditioned pipeline for single-image to 3D scene generation that jointly conforms to the object layout in input image and preserves instance fidelity. \name{} employs a hybrid instance generator that combines semantics-aware vector-set representation with detail-rich structured latent representation, yielding object geometries that are both semantically plausible and high-quality. Furthermore, \name{} enables the application of foundational pose estimation models in the scene generation task via a simple yet effective camera-conditioned scale-solving algorithm, to enforce scene-level coherence. Extensive experiments demonstrate that \name{} consistently generates high-fidelity camera-aligned compositional scenes, outperforming all state-of-the-art methods.

\end{abstract}

%% file: sec/1_intro.tex
\section{Introduction}

The rapid progress of foundation models has revolutionized numerous research domains, demonstrating impressive generalization capabilities in text-to-image synthesis\cite{ho2020ddpm,song2020ddim}, 3D generation\cite{liu2023syncdreamer,long2024wonder3d,hong2023lrm,tang2025lgm}, embodied intelligence\cite{kim2024openvlaopensourcevisionlanguageactionmodel,yu2023languagerewardsroboticskill,brohan2023rt2visionlanguageactionmodelstransfer}, and many others. Despite these successes, to achieve generalizability in single image-to-scene generation methods remains a challenge. Previous methods\cite{huang2025midi,lin2025partcrafter} often depend heavily on training with carefully curated, task-specific datasets, which inherently constrains their ability to generalize beyond the training distributions. Moreover, unlike large-scale 3D instance-level datasets\cite{deitke2023objaverse,deitke2024objaversexl}, curating diverse 3D scene-level datasets is significantly more difficult, resulting in limited data availability and reduced cross-dataset generalization. 

\begin{figure}
    \centering
    \includegraphics[width=0.48\textwidth]{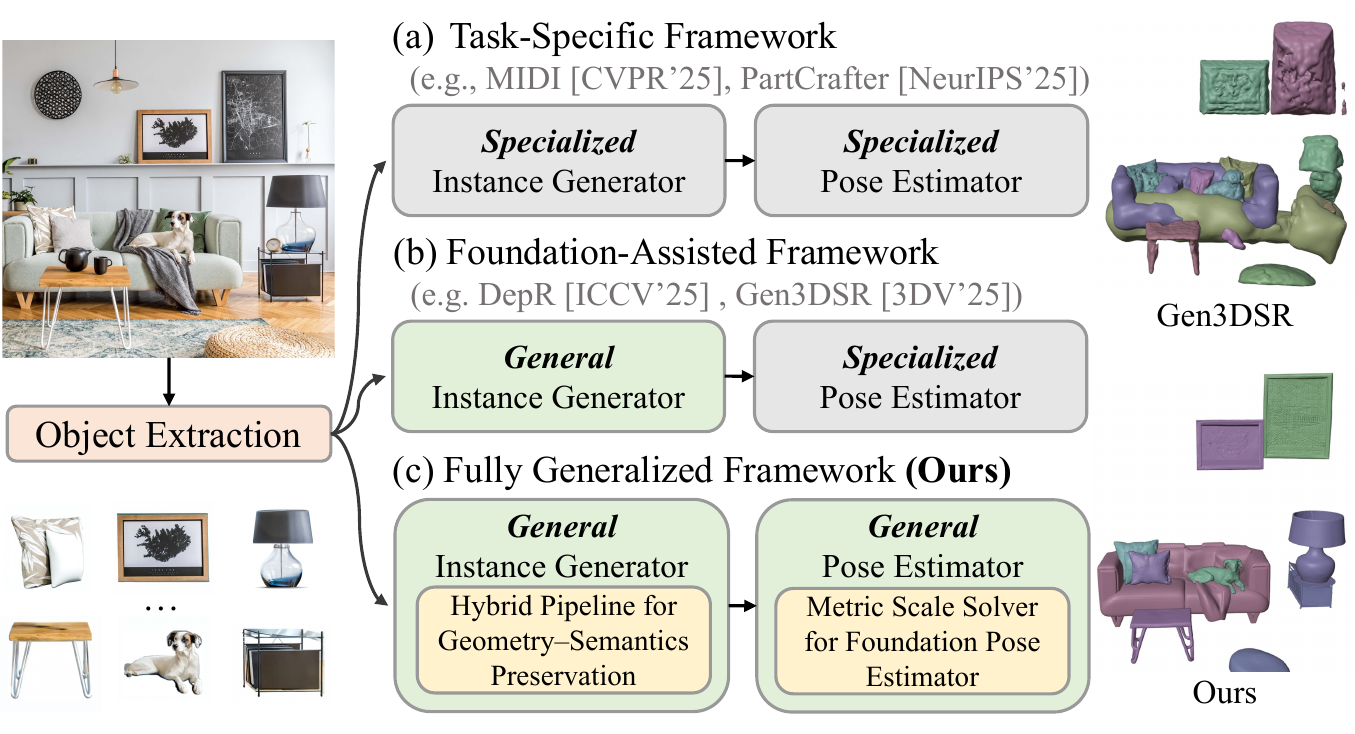}
    \caption{\name{} generates high-fidelity 3D scenes from single images in a zero-shot manner by addressing key challenges in applying 3D foundation models for scene generation. It combines (1) a hybrid instance generator to preserve instance semantics and geometric fidelity and (2) a foundation pose estimation model augmented with a novel metric-scale solving algorithm.}
    \label{fig:comparison}
\vspace{-10pt}
\end{figure}

Note that single-image to 3D scene generation is commonly formulated as a two-stage process comprising (1) instance generation and (2) pose estimation, and existing methods primarily differ in how these two stages are conducted, as illustrated in Fig.~\ref{fig:comparison}. Given the aforementioned limitation, several recent approaches~\cite{dogaru2024gen3dsr,zhou2024deepprior,zhao2025depr} integrate foundation models into single-image 3D scene generation pipelines, but only at selected stages rather than throughout the pipeline. They mostly utilize foundation models for instance generation, but employ lightweight, task-specific pose estimators or classical point-cloud registration techniques for subsequent pose estimation. These pose estimators frequently become the dominant source of error, and as a consequence, limiting the overall robustness and generalization capability of such methods, preventing them from generalizing reliably across diverse scenes and camera setups. To the best of our knowledge, none of existing methods has realized a fully foundation-model-driven pipeline, particularly for the pose estimation stage. This motivates the central research question of our work:

\emph{Can we achieve generalizable 3D scene generation from a 2D image with 3D foundation models only?}

In this research, we aim to answer this question by introducing a fully foundation model–powered pipeline for generalizable 3D scene generation. Specifically, we identify two key issues that impact the performance of foundation models in compositional scene generation: occlusion-induced errors during instance generation, where the generated instances deviate from the correct object semantics; and scale mismatch, where the object scales of generated meshes deviate from real-world scales.

To address occlusion-induced errors, we propose a hybrid instance generator that preserves object semantics using the vector latent set representation (VecSet) \cite{zhang20233dshape2vecset} while improving geometric fidelity with structured latents (SLAT)\cite{xiang2025structured3dlatentsscalable}. Specifically, we first leverage a massively pretrained VecSet-based object generator to produce coarse 3D assets. VecSet’s low-dimensional latent formulation captures semantic structure robustly—even under occlusion—thus preventing semantics from shifting toward incorrect object types, as shown in Fig. \ref{fig:ablation}. The coarse geometry is then refined by a generic detail-enhancing model built on SLAT representation and normal estimation\cite{ye2024stablenormal}, which produces detailed geometry without altering the underlying semantics. To resolve the scale mismatch problem in pose estimation, we propose a simple yet effective camera-conditioned scale solving algorithm that admits a closed-form linear solution. This algorithm is conditioned on camera parameters and produces metric scales across components, enabling foundation-scale pose estimation models to be used directly for compositional scene generation, without requiring specialized, task-specific pose estimation modules.

In summary, our main contributions are:

\begin{itemize}
    \item We propose \name, a fully generalized framework for zero-shot 3D scene generation from single images.
    \item We introduce a novel hybrid instance generation paradigm for both semantics-preserving and high-fidelity 3D object generation, together with a simple yet effective camera-conditioned algorithm with closed-form solution, to enable the application of foundation pose estimation models to the compositional scene generation task.
    \item We evaluate \name{} for generating diverse open-world 3D scenes that align with camera parameters, and show its superiority over the state-of-the-art zero-shot or training-dependent methods.
\end{itemize}

%% file: sec/2_related_works.tex
\section{Related Works}

\subsection{Single Image to 3D Scene Generation. }
Generating a set of posed 3D scenes given a single image remains a core and long-standing problem in computer vision. Depending on how to process the scene, current approaches in the literature can be broadly classified into two categories: feed-forward reconstruction techniques~\cite{nie2020total3d,zhang2021im3d,paschalidou2021atiss,dahnert2021panorecon,liu2022instpifu,gkioxari2022usl,zhang2023uni3d,chu2023buol,chen2024ssr}, and compositional generation frameworks~\cite{zhou2024deepprior,chen2024comboverse,han2024reparo,dogaru2024gen3dsr,tang2024diffuscene,huang2025midi,yao2025castcomponentaligned3dscene}.

\paragraph{Feed-Forward Scene Reconstruction. }
Feed-forward reconstruction techniques~\cite{nie2020total3d,zhang2021im3d,paschalidou2021atiss,dahnert2021panorecon,liu2022instpifu,gkioxari2022usl,zhang2023uni3d,chu2023buol,chen2024ssr} typically formulate scene reconstruction as a regression problem, exploiting spatial regularities of indoor environments and relying on 3D data to train networks in an end-to-end manner. They commonly adopt encoder-decoder architectures to predict scene geometry directly from a single input image. Although jointly inferring scene layout and object poses simplifies the training objective, the reconstruction quality is often limited by the scarcity and bias of annotated 3D scene datasets, leading to poor generalization on out-of-distribution objects. 
In the single-image setting considered in this work, scene reconstruction is conditioned on a single 2D view, so the network is strongly constrained only on visible pixels, while occluded and back-facing regions cannot be reliably reconstructed. As a result, such methods often produce incomplete or biased object geometries that do not faithfully capture the underlying scene.

\paragraph{Compositional Scene Generation.}
Recent compositional generation frameworks~\cite{zhou2024deepprior,chen2024comboverse,han2024reparo,dogaru2024gen3dsr} integrate foundation models for both image~\cite{kirillov2023sam,liu2023groundingdino,ren2024groundedsam,ramesh2022dalle2,nichol2022glide,saharia2022imagen,rombach2022ldm,podell2023sdxl} and 3D object~\cite{jun2023shape,eftekhar2021omnidata,zhang2024clay} domains to enhance scene reconstruction. These approaches generally follow a multi-stage procedure involving image segmentation~\cite{ren2024groundedsam}, object completion~\cite{rombach2022ldm}, per-object generation~\cite{jun2023shape,zhang2024clay}, and pose estimation~\cite{wen2024foundationpose}. Although they improve generalization by utilizing pre-trained models, none of them is able to apply foundational pose estimation abilities for scene generation, limiting the applicability and generalization of the overall pipelines. Our method tackles the problem by employing a foundation pose estimation model, which is made possible with a simple yet effective scale solving algorithm with closed-form solutions.

\subsection{Single Image to 3D Object Generation}
Recent advancements of 3D generation~\cite{liu2024one2345,liu2023syncdreamer,long2024wonder3d,hong2023lrm,tang2025lgm,huang2024epidiff,zhang2024clay,wu2024unique3d,li2024craftsman,wen2024ouroboros3d,xu2024instantmesh,voleti2025sv3d,wang2024crm,liu2024one2345++,wu2024direct3d,zhao2024michelangelo,roessle2024l3dg,wu2024blockfusion,meng2024lt3sd,liu2024part123,dong2025tela} has been largely driven by progress in diffusion models~\cite{ho2020ddpm,song2020ddim} and the creation of large-scale datasets~\cite{deitke2023objaverse,deitke2024objaversexl}. A number of image-to-3D object generation works~\cite{liu2023syncdreamer,long2024wonder3d,tang2025lgm,wen2024ouroboros3d,xu2024instantmesh,wang2024crm,voleti2025sv3d,huang2024mvadapter} generate 3D assets in two stages that first generates multi-view images and then reconstructs the 3D geometry from these views. These methods typically involve fine-tuning pre-trained image~\cite{rombach2022ldm,podell2023sdxl} or video~\cite{blattmann2023svd} diffusion models to produce coherent multi-view images, and subsequently recover the 3D shape with either large-scale reconstruction models~\cite{hong2023lrm,tang2025lgm,xu2024grm,zou2024tgs} or optimization-based techniques~\cite{wang2021neus}. There are also 3D-native techniques~\cite{zhang2024clay,li2024craftsman,wu2024direct3d,zhao2024michelangelo,li2025triposg} that concentrate on generating 3D-native geometries by training large-scale generative models, which often train a latent diffusion transformer (DiT)~\cite{peebles2023dit} over the latent space of a variational autoencoder~\cite{kingma2013vae}. Owing to the diversity and scale of datasets they are trained on, these models are capable of producing high-quality 3D geometries with remarkable generalizability. Our work is built upon these works, integrating both a low-dimensional VecSet-based object generation model for semantics preservation, and a high-dimensional SLAT-based one for geometry fidelity.

\subsection{Pose Estimation} 
Depending on the level of generalization, pose estimation techniques can be roughly divided into three categories: instance-level\cite{wang2019densefusion, peng2019pvnet, he2020pvn3d, Li2019CDPN, Zakharov2019DPOD}, category-level\cite{SGPA, DPDN, GPV-Pose, HS-Pose, IST-Net} and unseen object methods\cite{Gen6D, MegaPose, GigaPose, SAM-6D, FoundationPose}, in which a model is able to either: a) only estimate the pose of a specific object on which it is trained, b) estimate intra-class unseen instances if the class has been seen during training, or c) handle unseen object categories during training. In our setting we take advantage of the unseen object methods owing to the richness of objects in the input scenes. However, the foundation models for unseen object pose estimation are still deficient, where FoundationPose is a representative. In this work we bridge the gap between  FoundationPose and the compositional scene generation task to achieve accurate and generalizable pose estimation.

%% file: sec/3_methods.tex
\section{Methodology}
\begin{figure*}[!t]
    \centering
    \includegraphics[width=0.95\textwidth]{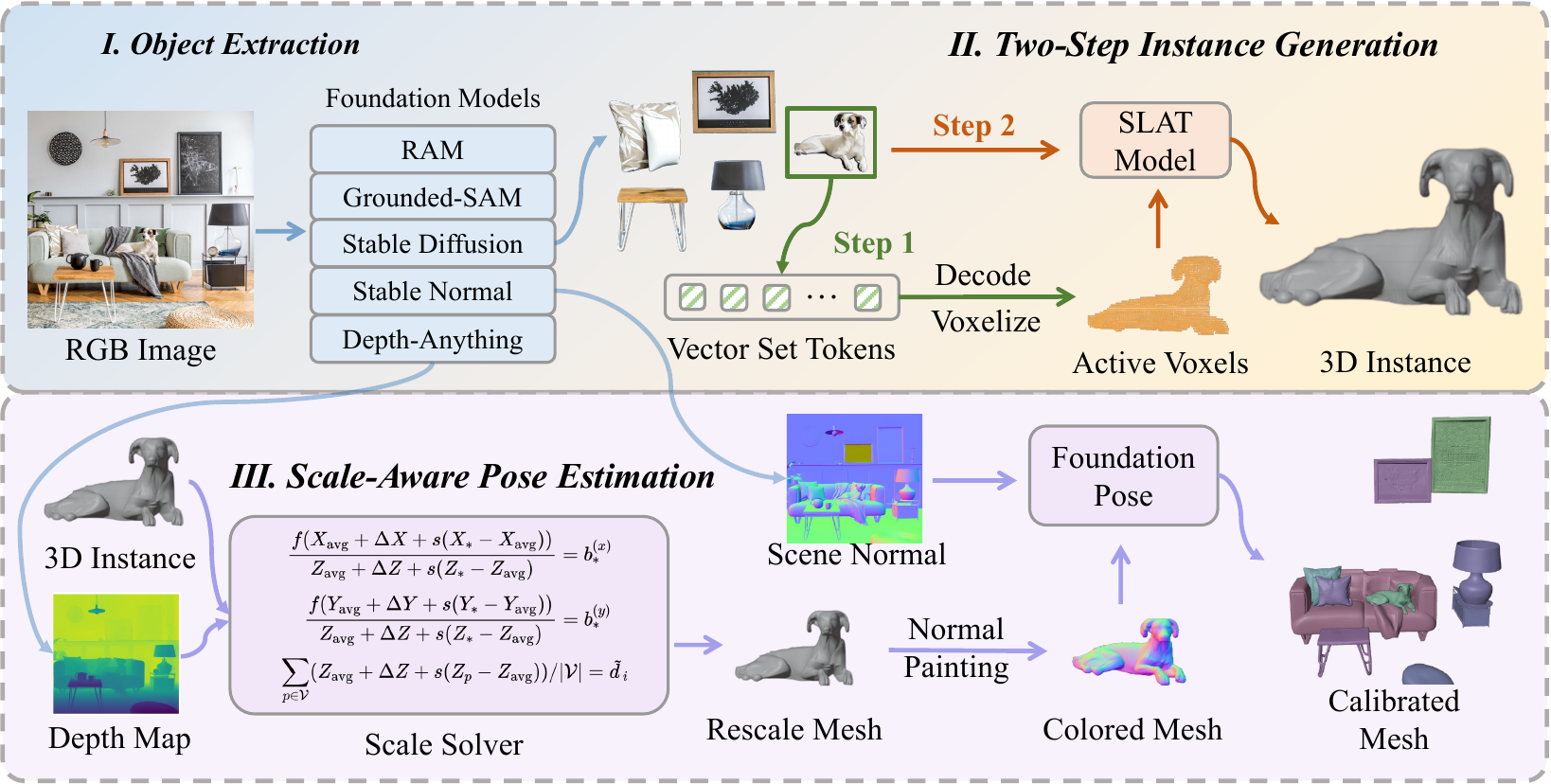}
    \caption{Illustration of \name{}, a zero-shot framework built entirely on foundation models for single-image to 3D scene generation. Our method delivers semantically accurate, high-fidelity geometry alongside precise per-object pose estimation. We first extract clean object instances through occlusion-aware segmentation and inpainting, providing reliable inputs for 3D generation. Next, we synergize a hybrid 3D foundation generation approach that melds vector latent-set (VecSet) and structured-latent (SLAT) models to synthesize high-quality, semantically consistent instances. Finally, to ensure coherent scene composition, we orchestrate a scale-solving algorithm with a foundational pose estimation model to resolve scale ambiguity and align generated objects with the input's spatial layout in a camera-conditioned manner.}
    \label{fig:pipeline}
\end{figure*}
The goal of \name{} is to develop a generalizable method that generates 3D scenes directly from RGB images, performing robustly across diverse real-world environments. To this end, we design a zero-shot pipeline built entirely on foundation models to achieve both semantics preservation together with enhanced geometry fidelity in instance generation, and improved accuracy in per-object pose estimation.
We first carefully segment and inpaint 2D object instances with occlusion awareness using 2D foundation models, providing clean inputs for subsequent 3D instance generation (Sec.~\ref{sec:2d-preprocessing}). To produce high-quality geometry and semantically accurate instances, we integrate VecSet and SLAT 3D foundation generation models for instance synthesis (Sec.~\ref{sec:object-generation}). Lastly, for coherent scene composition, we incorporate a foundational pose estimation model with a camera conditioned scale solver that resolves scale ambiguity inherent to 2D images, to align the generated objects with indicated spatial layout (Sec.~\ref{sec:pose-estimation}).
Together, these components form a coherent and generalizable framework capable of efficiently generating complex 3D scenes (see Fig.~\ref{fig:pipeline}).

\subsection{2D Preprocessing}\label{sec:2d-preprocessing}

\paragraph{Object Extraction.}
To prepare a 2D image $I$ for 3D instance generation, we first employ an open-set image tagging model, Recognize Anything~\cite{zhang2024recognize}, to obtain object labels $\{l_i\}_{i=1}^N$, where $N$ denotes the number of objects in $I$. For each label $l_i$, we extract the corresponding 2D bounding box $b_i$ and segmentation mask $m_i$ using the open-vocabulary object segmentation model Grounded-SAM~\cite{ren2024groundedsam}. We then remove background categories such as ``floor,'' ``room,'' and ``wall''. Using the remaining object masks, we extract the associated pixels to produce instance-level 2D object images $\{O_i\}_{i=1}^N$ for subsequent processing.

\paragraph{Occlusion-Aware 2D Inpainting.} Given that $\{O_i\}_{i=1}^N$ tends to suffer from object occlusion, before sending them to the 3D generation models we inpaint those occluded $O_i$, to prevent lose of information. We first employ the widely adopted Depth-Anything-v2\cite{yang2024depth}, together with previously estimated masks, to estimate per-object depths $\{D_i\}_{i=1}^N$. 
We determine the occlusion of each object instance $i$, by jointly considering its adjacent objects in the image and their depths. For each pair of adjacent objects, the instance with the smaller depth value is considered unoccluded, while the other is considered occluded and processed by our inpainting model to restore the occluded regions. We utilize Stable-Diffusion-XL-1.0-Inpainting-0.1~\cite{podell2023sdxl} due to its wide usage. %, with prompt ``a complete model of $l_i$, white background". 这句太细节了，放附录
The corresponding instance masks are updated with the inpainted results accordingly. Our depth-guided inpainting procedure distinguishes itself from previous approaches~\cite{han2024reparo, dogaru2024gen3dsr} in that it avoids unnecessary inpainting, thereby preserving the fidelity of the original object images as much as possible.

%%%%%%%%%%%%%%%%%%%%%%%%%%

\subsection{Object Generation}\label{sec:object-generation}

Recent works\cite{xiang2025structured3dlatentsscalable,ye2025hi3dgen,wu2025direct3ds2gigascale3dgeneration} based on structured latent representation (SLAT) have shown excellent generation performance on real-world image-based 3D instance generation. Unfortunately, in our experiments (see Fig. \ref{fig:ablation}), SLAT-based models alone exhibit suboptimal performance, as they tend to focus on local geometry details, instead of overall object semantics. For instance, given an image of a framed painting, SLAT-based methods such as Stable3DGen\cite{ye2025hi3dgen} incorrectly interpret it as the side of a delicate aquarium with aquatic plants. Therefore, inspired by Ultra3D\cite{chen2025ultra3d}, we employ a two-stage object generation pipeline.

First, to preserve object semantics, we generate a coarse 3D object using Step1X3D\cite{li2025step1x}, a vector latent-set (VecSet)–based object generator trained on large-scale instance data. Due to its low-dimentional nature, the VecSet representation provides high-level semantic guidance, encouraging the model to capture the global semantics of the object. However, in contrast to SLAT-based methods, VecSet-based ones alone generally fall short of geometry fidelity, as shown in Fig. \ref{fig:ablation}. To leverage the strengths of both, we voxelize the output of Step1X3D into indices of active voxels, and then employ the widely used Stable3DGen\cite{ye2025hi3dgen} to predict the corresponding structured latents, which are then populated into the decoder for fine-grained object meshes $\{M_i\}_{i=1}^N$. Although our two-stage generation procedure is reminiscent of that of Ultra3D, our approach distinguishes itself in that it employs foundation models, instead of specifically trained ones, in the pipeline. Furthermore, our design purpose differs, as we use VecSet models primarily for semantics extraction rather than resolution enhancement, thus adopting no voxel super-resolution techniques.

\subsection{Scale-Aware Pose Estimation}\label{sec:pose-estimation}

Given the scene image $I$, camera parameters $c$, and generated meshes $\{M_i\}_{i=1}^N$, a central challenge is to recover per-object transformations $(s_i, R_i, t_i)$ to align the meshes with input spatial layouts. Noticeably, no prior work has successfully applied a foundation pose estimation model to scene generation. Existing methods either train a scale-aware pose estimation module from scratch \cite{huang2025midi}, or resort to time-consuming optimization on 3D discrepancy\cite{zhou2024deepprior}, which, in practice (see Tab. \ref{tab:comp_metrics}), often degrade on out-of-domain images and produce unstable and inaccurate pose estimations.

A naive approach is to apply open-set foundation pose estimation models, such as FoundationPose~\cite{wen2024foundationpose}, to our problem. However, this cannot be done directly due to scale mismatch between the generated meshes and the target scene. In particular, FoundationPose only estimates the rotation $R$ and translation $t$ of an object while assuming a known scale $s$. Should this assumption be violated, pose estimation accuracy degrades substantially (see Tab.~\ref{tab:ablation_on_pose_est}). 
In our setting, the instance-generation foundation models are trained on objects with diverse and unconstrained physical sizes, thus, the scales of the generated meshes $M_i$ are generally inconsistent with the metric scale implied in the input image. This mismatch makes naive application of foundation pose estimators infeasible for accurate 3D scene generation.

To incorporate FoundationPose into our pose estimation procedure, we propose a scale estimation algorithm, which together with FoundationPose constitutes the overall pose estimation module of our scene generation pipeline, as illustrated in Fig. \ref{fig:pipeline}.

\paragraph{Solving for Scale.} We denote the translation and scale to be solved by $t := (\Delta X, \Delta Y, \Delta Z)$ and $s$, respectively. Our derivation is based on the standard pinhole camera projection model, $\frac{f}{Z} X = x$, where $X$ is the $x$-coordinate of a 3D point in camera space, $Z$ is its depth, $f$ is the focal length, and $x$ is the corresponding $x$-coordinate on the image plane. We formulate the constraints for the algorithm in both 2D and 3D spaces.

\begin{itemize}
    \item \textbf{Constraint 1}: After rescaling and translation, the 2D bounding box of $M_i$ should align with $b_i$.
\end{itemize}

To enforce the constraint, we first rasterize $M_i$, to get the uppermost, lowermost, leftmost and rightmost 2D points $(x_*, y_*)$, and their 3D counterparts $(X_*, Y_*, Z_*)$, where $*$ can be one of ``upper", ``lower", ``left" or ``right". 
% We denote the four 2D points and corresponding 3D points as $(x_*, y_*)$ and $(X_*, Y_*, Z_*)$ respectively, where $*$ can be one of ``upper", ``lower", ``left" or ``right".

Next, we compute the 2D bounding box of the transformed mesh. When a point $(X, Y, Z)$ is transformed by translation $(\Delta X, \Delta Y, \Delta Z)$ and scaling $s$, its new location becomes:
\begin{equation}\label{eq:translate_and_scale_XYZ}
\begin{aligned}
&\mathcal{T}_p((X, Y, Z), (\Delta X, \Delta Y, \Delta Z, s))=\\
&(X_{\text{avg}} + \Delta X + s (X - X_{\text{avg}}),\\ 
&Y_{\text{avg}} + \Delta Y + s (Y - Y_{\text{avg}}),\\
&Z_{\text{avg}} + \Delta Z + s (Z - Z_{\text{avg}})),
\end{aligned}
\end{equation}
where $X_{\text{avg}}$, $Y_{\text{avg}}$, and $Z_{\text{avg}}$ are the means of the $X$, $Y$, and $Z$ coordinates of mesh $M_i$, respectively, and $\mathcal{T}_p$ denotes the pointwise linear transformation. To make the computation of the bounding boxes of the transformed meshes tractable, we approximate that the leftmost, rightmost, uppermost, and lowermost visible 3D points of a mesh remain to be the extremal visible points after transformation.

With this approximation, substituting Eq.~\ref{eq:translate_and_scale_XYZ} into the pinhole camera projection model yields the 2D coordinate of the leftmost visible point of the transformed mesh, which should match the $x$-coordinate of the leftmost edge of $b_i$, namely:
\begin{algorithm}[!t]
\caption{Solving for object-wise scale and translation with closed-form solution. Omit superscript i for brevity}\label{alg:linear}
\begin{algorithmic}
\Require $X_{\text{avg}}$, $Y_{\text{avg}}$, $Z_{\text{avg}}$, $\{X_*\}$,$\{Y_*\}$,$\{Z_*\}$,$\{b_*^{(x)}\}$, $\{b_*^{(y)}\}$, $\forall *\in \{\text{left}, \text{right}, \text{upper}, \text{lower}\}$, $\tilde{d}$, $f$, $Z$ \Comment{$Z$ is the object-wise depth map}
\State $dX_{*}, dY_{*}, dZ_{*} \gets X_{*}-X_{\text{avg}},Y_{*}-Y_{\text{avg}},Z_{*}-Z_{\text{avg}}$

\State $dZ \gets Z - Z_{\text{avg}}$

\State $A \gets \begin{bmatrix}
f & 0 & -b_\text{left}^{(x)} & f \cdot dX_\text{left} - b_\text{left}^{(x)} \cdot dZ_\text{left} \\
f & 0 & -b_\text{right}^{(x)} & f \cdot dX_\text{right} - b_\text{right}^{(x)} \cdot dZ_\text{right} \\
0 & f & -b_\text{upper}^{(y)} & f \cdot dY_\text{upper} - b_\text{upper}^{(y)} \cdot dZ_\text{upper} \\
0 & f & -b_\text{lower}^{(y)} & f \cdot dY_\text{lower} - b_\text{lower}^{(y)} \cdot dZ_\text{lower} \\
0 & 0 & 1 & \frac{\sum dZ}{\sum \mathbf{1}[ Z \neq 0]}
\end{bmatrix}$
\State $B \gets \begin{bmatrix}
b_\text{left}^{(x)} \cdot Z_{\text{avg}} - f \cdot X_{\text{avg}} \\
b_\text{right}^{(x)} \cdot Z_{\text{avg}} - f \cdot X_{\text{avg}} \\
b_\text{upper}^{(y)} \cdot Z_{\text{avg}} - f \cdot Y_{\text{avg}} \\
b_\text{lower}^{(y)} \cdot Z_{\text{avg}} - f \cdot Y_{\text{avg}} \\
\tilde{d} - Z_{\text{avg}}
\end{bmatrix}$
\State Solving $AX=B$, where $X=\begin{bmatrix}
\Delta X\\
\Delta Y\\
\Delta Z\\
s
\end{bmatrix}$ %\Comment{Done with torch.linalg.lstsq}
\end{algorithmic}
\end{algorithm}

\begin{equation}\label{eq:bounding-box-no-i}
    \frac{f(X_{\text{avg}} + \Delta X + s (X_\text{left} - X_{\text{avg}}))}{Z_{\text{avg}} + \Delta Z + s (Z_\text{left} - Z_{\text{avg}})} = b_\text{left}^{(x)}
\end{equation}

where the superscript $(i)$ is omitted for simplicity, and $b_{left}^{(x)}$ represents the $x$-coordinate of the leftmost edge in $b_i$. For brevity, we only show the equation for the leftmost visible points. The equations for the other visible points follow analogously, yielding four equations in total. Note that these equations are linear with respect to $t_i$ and $s_i$.

Due to depth–scale ambiguity, the equations defined in Eq.~\ref{eq:bounding-box-no-i} admit an infinite number of solutions for $t$ and $s$. To constrain the solution space, we define a second constraint:

\begin{itemize}
    \item \textbf{Constraint 2}: After transformation, the average depth of the visible region of an object matches that of the same object in $I$.
\end{itemize}
To obtain the depth of $I$, we run Depth-Anything-v2~\cite{yang2024depth} to estimate a depth map. We then compute the average depth of the $i$-th object's visible region, denoted by $\tilde{d}_i$. This yields another equation:
\begin{equation}\label{eq:depth}
\sum_{p \in \mathcal{V}} \left(Z_{\text{avg}} + \Delta Z + s (Z_p - Z_{\text{avg}})\right)/|\mathcal{V}| = \tilde{d}_i
\end{equation}
where the numerator is the depth of all visible points of the transformed mesh, and the denominator $\mathcal{V}$ denotes the set of all visible points.

We jointly solve the five linear equations defined in Eq.~\ref{eq:bounding-box-no-i} and Eq.~\ref{eq:depth} for $s_i$ and $t_i$. Since these equations are linear with respect to $s$ and $t$, they can be solved efficiently using standard linear equation solvers within milliseconds. Due to the approximation introduced in \emph{Constraint 1}, the resulting $t_i$ and $s_i$ are not perfectly accurate. We therefore iteratively solve the equations multiple times to refine $s_i$ and $t_i$. In our experiments, we find that four iterations are sufficient. The algorithm for one step of scale solving is shown in Alg.~\ref{alg:linear}.

\paragraph{Mesh Texturing for Accurate Pose Estimation.}With $\{s_i\}_{i=1}^N$ and $\{t_i\}_{i=1}^N$, applying FoundationPose becomes feasible. However, a key challenge remains: FoundationPose assumes textured meshes as input, whereas generating consistent object textures from a single RGB image is inherently difficult. In our preliminary experiments, we employed a state-of-the-art mesh inpainter~\cite{huang2024mvadapter} and observed that the inpainting quality was limited, which in turn severely degraded pose estimation accuracy.
To address this issue, we texture the generated meshes using vertex normals and estimate the normal map of $I$, denoted by $I_{\text{normal}}$, using StableNormal~\cite{ye2024stablenormal}. We then invoke FoundationPose on $I_{\text{normal}}$ together with the normal-textured meshes to obtain estimates of $R_i$. Given $R_i$, we solve the five linear equations to obtain refined $s_i$ and $t_i$, yielding the final transformed meshes. By encoding geometric information into the texture in the form of normal maps, this mechanism encourages FoundationPose to focus on aligning geometric details rather than appearance textures. %The experiments reported in Tab.~\ref{tab:ablation_on_pose_est} validate the effectiveness of this design.

\paragraph{Camera-Conditioned Generation.}
Unlike recent training-dependent scene generation methods~\cite{huang2025midi,lin2025partcrafter,zhao2025depr}, CC-FMO is camera-conditioned: the focal length and other camera parameters naturally enter the formulation of our constraints and are preserved throughout. This is an important feature because monocular images suffer from well-known scale ambiguity: different combinations of object scale, depth, and focal length can produce identical 2D projections. By enforcing consistency with camera parameters in our constraints, we effectively restrict the solution space and recover object poses and scales that are geometrically meaningful rather than only image-consistent. 

Moreover, as demonstrated in our experiments (see Tab. ~\ref{tab:comp_metrics}), our model can zero-shot adapt to scenes with different camera parameters at test time simply by updating the intrinsics used in the projection equations. This feature stands in contrast to prior works~\cite{huang2025midi,lin2025partcrafter}, which typically assume fixed intrinsics and absorb such prior information into learned parameters, limiting the ability to generalize across cameras or to produce metrically calibrated 3D scenes.

%% file: sec/4_experimentation.tex
\section{Experiments}

\begin{table}[t]\small
    \centering
    \setlength{\tabcolsep}{3pt}
    \caption{Quantitative results on 3D-FRONT~\cite{fu20213dfront} measured by scene-level Chamfer Distance (CD-S) and F-Score (F-Score-S), object-level Chamfer Distance (CD-O) and F-Score (F-Score-O), and object bounding box Volume IoU (IoU-B). Colors indicate whether methods are \colorbox{red!20}{training-dependent} or \colorbox{green!20}{zero-shot}. Best and second-best results are in bold and underlined, respectively. CD and FScore are shown in percentage.}
    \label{tab:comp_metrics}
    \resizebox{\linewidth}{!}{%
    \begin{tabular}{l|ccccc}
    \toprule
    Method  & CD-S$\downarrow$ & FScore-S$\uparrow$ & CD-O$\downarrow$ & FScore-O$\uparrow$ & IoU-B$\uparrow$  \\
    \midrule
    \colorbox{red!20}{DepR}~\cite{zhao2025depr} & 10.71 & 39.81 & 22.24 & 39.51 & 0.142  \\
    \colorbox{red!20}{PartCrafter}~\cite{lin2025partcrafter} & 10.82 & 47.19 & 3.201 & 80.10 & 0.294  \\
    \colorbox{red!20}{MIDI}~\cite{huang2025midi} & \underline{2.241} & \underline{75.81} & \underline{0.802} & \underline{89.42} & \underline{0.527} \\
    \colorbox{green!20}{Gen3DSR}~\cite{dogaru2024gen3dsr} & 22.81 & 26.78 & 10.78 & 54.95 & 0.113  \\
    \colorbox{green!20}{DPA*}~\cite{zhou2024deepprior} & 12.47 & 37.71 & 4.054 & 70.29 & 0.137    \\
    \colorbox{green!20}{Ours} & \textbf{1.805} & \textbf{89.60} & \textbf{0.738} & \textbf{91.00} & \textbf{0.670} \\
    \bottomrule
    \end{tabular}
    }
\end{table}

\begin{figure*}
    \centering
    \includegraphics[width=0.98\linewidth]{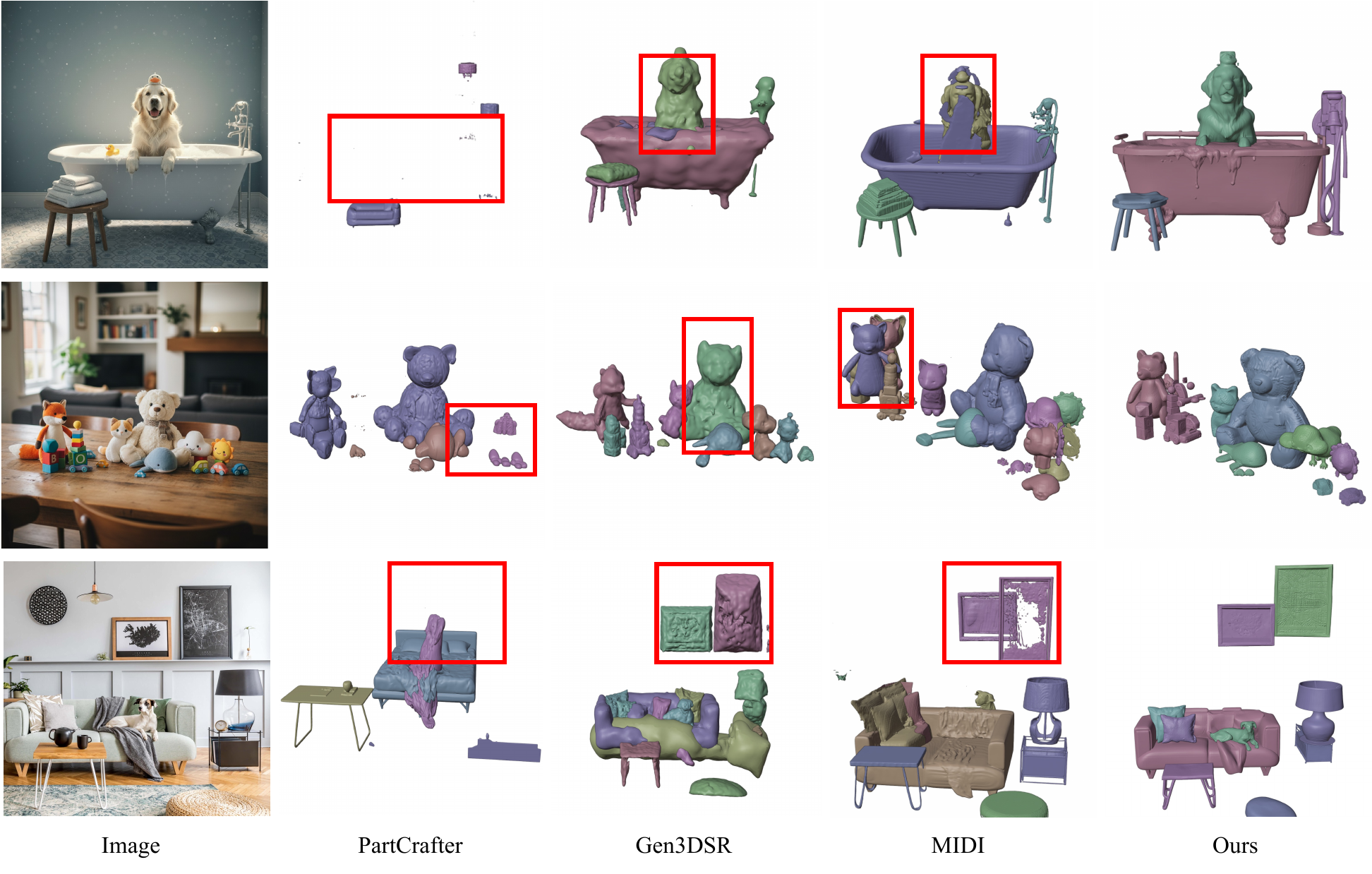}
    \caption{Qualitative results. \name{} demonstrates strong performance in this challenging scenario. Compared to baseline methods, our model produces meshes with superior geometric fidelity while maintaining object layouts that closely align with the input images. Refer to our supplementary materials for more qualitative comparisons.}
    \label{fig:qualitative}
\end{figure*}

\subsection{Experimental Setup}
For fair and consistent comparisons, we follow the evaluation protocol of MIDI~\cite{huang2025midi}. All random seeds are fixed to 0 for reproducibility. The details are as follows:

\paragraph{Implementation Details.} Recall that our pipeline operates in a zero-shot setting. Object detection and segmentation are conducted with Grounded-SAM~\cite{ren2024groundedsam}, where Recognize Anything~\cite{zhang2024recognize} is first used to obtain object category labels. We adopt Stable-Diffusion-XL-1.0-Inpainting-0.1~\cite{podell2023sdxl} as our 2D inpainting model, followed by SAM~\cite{kirillov2023segment} to retrieve segmentation masks on the inpainted images. 

Object generation is performed using Step1X-3D~\cite{li2025step1x} and Stable3DGen~\cite{ye2025hi3dgen}, where the latter relies on StableNormal~\cite{ye2024stablenormal} for normal estimation. FoundationPose~\cite{wen2024foundationpose} estimates per-object poses given the estimated scene-level normal map and per-object normal maps from StableNormal, together with a scene depth map estimated by Depth-Anything-v2~\cite{yang2024depth}. The linear systems defined in Eq.~\ref{eq:bounding-box-no-i} and Eq.~\ref{eq:depth} are solved using the \texttt{linalg.lstsq} solver in PyTorch, and rasterization is implemented with PyTorch3D.

\paragraph{Datasets.} 
For quantitative evaluation, we adopt the MIDI test partition of 3D-FRONT~\cite{huang2025midi}. However, we observe that this partition contains samples that are ill-posed for scene generation evaluation. For example, some views place the cameras so far from the scene that the rendered content occupies only tiny, indistinguishable regions of the image, while others produce almost entirely black images due to severe occlusion by large foreground objects. We manually filter out such degenerate cases and evaluate all models on the remaining samples, which comprise more than 500 scenes.
In contrast to MIDI, our rendering setup randomizes the camera field of view between 20 and 60 degrees, and the camera position is adjusted to ensure that the scene occupies a substantial portion of the rendered image. The camera coordinate axes are aligned with the world coordinate axes, like MIDI. Following MIDI, we also compute ground-truth depth maps and normal maps during rendering, which are used in place of the outputs of Depth-Anything-v2 and StableNormal during quantitative evaluation.

It is worth noting that the real-world images in our evaluation are substantially more complex than those used in prior work~\cite{huang2025midi,lin2025partcrafter,zhao2025depr}. For example, Fig.~\ref{fig:qualitative} illustrates that our model can handle complex indoor scenes.

\paragraph{Baselines.}
We compare CC-FMO with single image-based compositional scene generation models, including MIDI\cite{huang2025midi}, DPA\cite{zhou2024deepprior}, Gen3DSR\cite{dogaru2024gen3dsr}, PartCrafter\cite{lin2025partcrafter} and DepR\cite{zhao2025depr}. As DPA is also a zero-shot pipeline with foundation instance generator more outdated than ours, we replace its instance generator with ours to make fair comparisons. Thus evaluation difference between DPA and CC-FMO results from our effectiveness in pose estimation. For methods which are dependent on segmentations, like MIDI, we provide them with the segmentation masks same to ours to eliminate the effects from these irrelevant factors. Also the camera parameters are set as the same across all the methods. These baselines adopt various recipes for generation and training. The comparison between our model and them thouroughly testifies the effectiveness of our model design.

\paragraph{Metrics.}

Following MIDI~\cite{huang2025midi}, we adopt the widely used Chamfer Distance (CD) and F-Score (with the default threshold of 0.1) to evaluate scene-level performance. In addition, we compute \emph{object-level} Chamfer Distances and F-Scores between each generated object and its ground-truth counterpart to assess geometric fidelity at the object level. 

We further calculate the Volumetric Intersection over Union (Volume IoU) between the bounding boxes of generated objects and those of the ground truth to evaluate the accuracy of the estimated object layout and spatial consistency.

\subsection{Quantitative Experiments}
As shown in Tab.~\ref{tab:comp_metrics}, our zero-shot pipeline outperforms all baselines on all scene-level metrics, which is particularly noteworthy given that MIDI is trained directly on 3D-FRONT. Since our evaluation dataset is rendered with different camera parameters and offsets, we attribute this superior performance over dataset-specific models to the camera-conditioned design of CC-FMO. In particular, training-dependent methods such as MIDI implicitly encode the training-time camera configuration in their model parameters. When the camera parameters at test time differ from those seen during training, their performance degrades substantially, as they tend to produce erroneous inter-object distances due to compression or expansion of depth of view across camera setups.

Moreover, CC-FMO achieves the best performance on all object-level metrics, namely CD-O and FScore-O. We attribute this advantage to the generalizable instance generation capability and occlusion robustness afforded by our carefully designed instance generation pipeline.

\subsection{Qualitative Experiments}
We qualitatively compare CC-FMO with baseline models on real-world images in Fig.~\ref{fig:qualitative}. The evaluated scenes can contain around 10 objects. As shown in the figure, pipelines powered by foundation models produce scenes with smoother and more coherent geometry, highlighting the effectiveness of foundation models for this task. In contrast, methods trained on specific datasets, such as MIDI, tend to generate meshes with noticeable geometric discontinuities, as illustrated in the bottom example. We hypothesize that this degradation arises from distributional shifts between real-world objects and the curated training datasets. These observations support the use of foundation models as the core of our pipeline. Compared to baselines, CC-FMO produces meshes with the better geometric quality, and the resulting object layouts aligns with the input images.

\begin{table}[t]\small
    \centering
    \setlength{\tabcolsep}{3pt}
    \caption{Ablation study of our hybrid instance generator. Quantitatively evaluated on 3D-FRONT~\cite{fu20213dfront} using object-level Chamfer Distance (CD-O) and F-Score (F-Score-O), shown in percentage. }
    \label{tab:ablation_on_instance_gen}
    \begin{tabular}{cc|cc}
    \toprule
    VecSet & SLAT & CD-O$\downarrow$ & FScore-O$\uparrow$\\
    \midrule
    & \checkmark & 1.369 & 81.06 \\
    \checkmark &  & 1.166 & 78.66 \\
    \checkmark & \checkmark & \textbf{0.756} & \textbf{89.11} \\
    \bottomrule
    \end{tabular}
\end{table}

\begin{table}[t]\small
    \centering
    \setlength{\tabcolsep}{3pt}
    \caption{Ablation study of our pose estimation method. Quantitatively evaluated on 3D-FRONT~\cite{fu20213dfront} using scene-level Chamfer Distance (CD-S), F-Score (F-Score-S), and object bounding box Volume IoU (IoU-B).  CD and FScore are shown in percentage.}
    \label{tab:ablation_on_pose_est}
    \begin{tabular}{ccc|ccc}
    \toprule
Scale    & Foundation- & Normal   & \multirow{2}{*}{CD-S$\downarrow$} & \multirow{2}{*}{FScore-S$\uparrow$} & \multirow{2}{*}{IoU-B$\uparrow$} \\
Solving  & Pose        & Painting &                                   &                                     &                                   \\
    \midrule
     & \checkmark & \checkmark & 16.40 & 32.56 & 0.351 \\
    \checkmark &  & \checkmark & 3.159 & 81.26 & 0.555 \\
    \checkmark & \checkmark &  & 2.714 & 83.22 & 0.569 \\
    \checkmark & \checkmark & \checkmark & \textbf{1.923} & \textbf{87.20} & \textbf{0.629} \\
    \bottomrule
    \end{tabular}
\end{table}

\subsection{Ablation Study}

\begin{figure}[t]
    \centering
    \includegraphics[width=\linewidth]{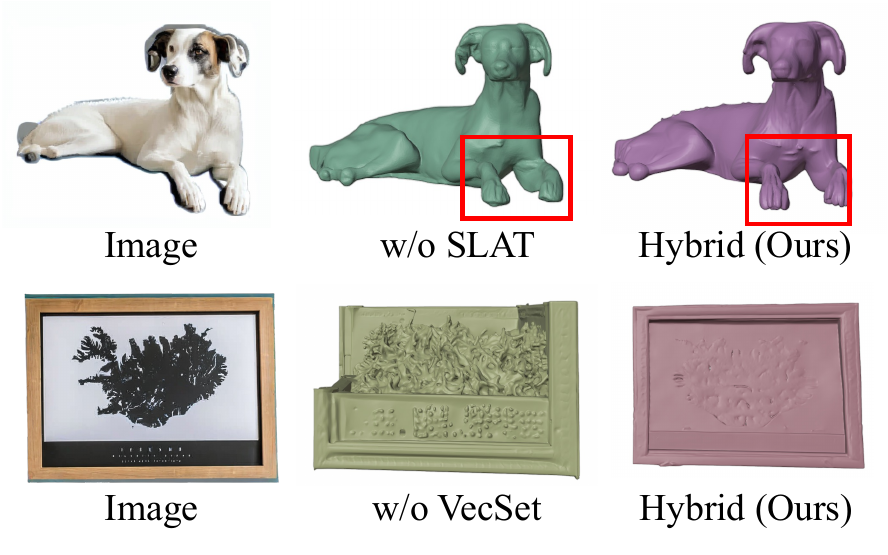}
    \caption{Ablation visualization. Removing the structured latent component (SLAT) leads to degraded geometric quality, while omitting the vector set (VecSet) results in reduced semantic accuracy. Our instance generator effectively preserves semantic accuracy and simultaneously produces detailed geometry.}
    \label{fig:ablation}
\end{figure}

\paragraph{Hybrid Instance Generation.}
As shown in Tab.~\ref{tab:ablation_on_instance_gen} and Fig.~\ref{fig:ablation}, ablating the proposed hybrid instance generation scheme yields objects with either incorrect semantics or noticeably degraded geometric details. In contrast, our instance generator strikes a balance between semantics preservation and geometric fidelity. Interestingly, although SLAT models produce instances with higher geometric fidelity when provided with unoccluded image inputs, removing SLAT from our instance generation pipeline leads to a smaller performance drop than removing the VecSet model. This observation highlights the crucial role of semantics preservation, particularly in occlusion-heavy scenarios.

\paragraph{Scale-Aware Pose Estimation.}
In Tab.~\ref{tab:ablation_on_pose_est}, we compare pose estimation performance with and without the proposed closed-form scale solving algorithm and FoundationPose. Due to the scale priors implicitly learned during the training of instance generation models, FoundationPose cannot be directly applied without explicit scale estimation. 

We further evaluate CC-FMO with the proposed normal-painting technique. The results demonstrate that this component is essential for accurate pose estimation. By encoding geometric details into the texture as normal maps, the method encourages FoundationPose to prioritize alignment of geometric structure over appearance-based textures.
\vspace{5pt}

%% file: sec/5_Conclusion.tex
\section{Future Work and Conclusion}

Building upon our work, future works can extend the input modalities for instance generation, instead of considering only images as inputs, which may fall short in fidelity-sensitive scenarios, due to scale-depth-focal ambiguity. 
We propose CC-FMO, a compositional scene generation framework with end-to-end foundation models. With proposed CC-FMO, we demonstrate that scene-level finetuning is not strictly necessary to achieve high-fidelity compositional scene generation. We identify key challenges specific to compositional scene generation that hinder the direct application of foundation models, and then address them through carefully designed model integration and a novel camera-aware, metric pose estimation algorithm. Our experiments show that CC-FMO generates high-fidelity scenes in a camera-conditioned manner. Despite being training-free, CC-FMO attains fidelity comparable to, or even surpassing, that of methods requiring task-specific training, highlighting the effectiveness of our design. We believe that our approach can benefit downstream applications that demand precise 3D scene generation.

%% file: sec/X_suppl.tex
\clearpage
\setcounter{page}{1}
\maketitlesupplementary

\section{More Implementation Details}

It is worth mentioning that $\{b_i\}_{i=1}^N$, $\{m_i\}_{i=1}^N$ and $\{l_i\}_{i=1}^N$ are paired, which enables us to filter out the object segmentations that do not stand for object instances at all, such as those with labels ``floor", ``room", or ``wall". 
To facilitate the subsequent object generation procedure, we enlarge the object regions in $\{I_i:=I*m_i\}_{i=1}^N$ by centering and normalizing, such that the non-zero region occupies $\alpha=60\%$ of the max width or height of the image, where the value is determined from small-scale experiments. The inpainting is done with prompt ``a complete model of $l_i$, white background". For all the models we align their generations with ground-truth ones by FilterReg\cite{gao2019filterregrobustefficientprobabilistic} during testing such that their orientations or offsets won't affect the scores. 

\section{Rationale for Baseline Choice}
The selected baselines span the spectrum between specifically-trained and foundation-assisted pipelines:
\begin{itemize}
    \item DepR is specifically trained. It utilizes depth information throught both training and inference, where the depth is provided by a foundation model.
    \item PartCrafter is specifically trained, which generates 3D instances in part-level with a 3D-native DiT\cite{Peebles2022DiT}. It is also able to be applied on scene generation by taking object instances as ``parts''.
    \item MIDI-3D is specifically trained, featured by compatibility with occluded object image inputs and a cross-instance attention mechanism to incorporate global context during scene generation.
    \item DPA is zero-shot. Whereas it generates 3D objects with foundation models, it calibrates generated objects by minimizing Chamfer Distances between point clouds with RANSAC-like algorithm.
    \item Gen3DSR is zero-shot. It decomposes scene generation as subtasks such as entity segmentation, depth estimation, etc. Each subtask is tackled via a different foundation model.
\end{itemize}
These baselines adopt various recipes for generation and training. The comparison between our model and them thouroughly testifies the effectiveness of our model design.

\section{More Qualitative Comparisons}
\begin{figure*}
    \centering
    \includegraphics[width=0.8\linewidth]{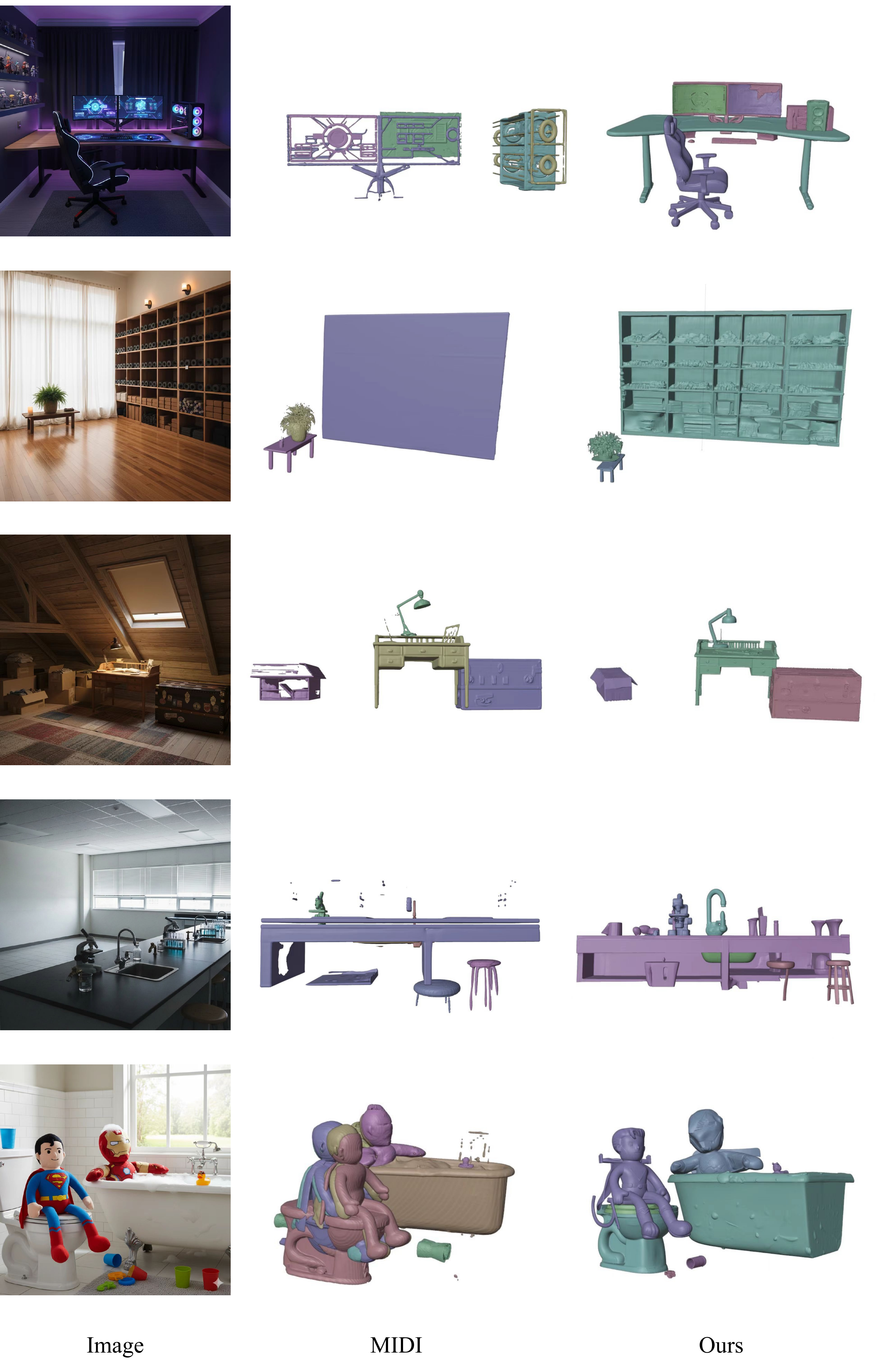}
    \caption{Qualitative comparison between CC-FMO and MIDI. Note that the segmentation masks are the same.}
    \label{fig:More_comparison}
\end{figure*}

In Fig. \ref{fig:More_comparison}, we provide more qualitative comparisons between our model and MIDI, the strongest baseline. The cases contain examples with various lights, backgrounds, numbers of objects, and scenes. Especially, the cameras are not aligned with the coordinates of objects, which makes the cases challenging.

